\pdfoutput=1
\documentclass[11pt]{article}

\usepackage[final]{acl}

\usepackage{booktabs}
\usepackage{times}
\usepackage{latexsym}

\usepackage[T1]{fontenc}
\usepackage[utf8]{inputenc}

\usepackage{microtype}

\usepackage{inconsolata}

\usepackage{graphicx}

\usepackage{amsmath}
\usepackage{afterpage}
\usepackage{graphicx}
\usepackage{xcolor}
\usepackage{array}

\title{keepitsimple at SemEval-2025 Task 3: LLM-Uncertainty based Approach for Multilingual Hallucination Span Detection}

\author{Saketh Reddy Vemula \\
  IIIT Hyderabad \\
  \texttt{saketh.vemula@research.iiit.ac.in} \\\And
  Parameswari Krishnamurthy \\
  IIIT Hyderabad \\
  \texttt{param.krishna@iiit.ac.in} \\}

\begin{document}
\maketitle
\begin{abstract}
Identification of hallucination spans in black-box language model generated text is essential for applications in the real world. A recent attempt at this direction is SemEval-2025 Task 3, Mu-SHROOM—a Multilingual Shared Task on Hallucinations and Related Observable Over-generation Errors. In this work, we present our solution to this problem, which capitalizes on the variability of stochastically-sampled responses in order to identify hallucinated spans. Our hypothesis is that if a language model is certain of a fact, its sampled responses will be uniform, while hallucinated facts will yield different and conflicting results. We measure this divergence through entropy-based analysis, allowing for accurate identification of hallucinated segments. Our method is not dependent on additional training and hence is cost-effective and adaptable. In addition, we conduct extensive hyperparameter tuning and perform error analysis, giving us crucial insights into model behavior.\footnote{The code is available at \url{https://github.com/SakethReddyVemula/semeval-2025_Mu-SHROOM}}
\end{abstract}

\section{Introduction}

Hallucination is a situation where Large Language Models (LLMs) produce outputs that are inconsistent with real-world facts or unverifiable, posing challenges to the trustworthiness of AI systems \citep{Survey_Hallucination}. Hallucination Detection is the process of identifying such sections of text where a model generates content that is untrue, misleading, or unverifiable by any source. As LLMs are used to generate massive texts in all applications, it is essential to make sure their output is accurate \citep{Foundation-Models}. Undetected hallucinations can propagate misinformation, lower confidence in AI systems, and have severe implications in applications such as healthcare and law. Identification of particular spans of hallucinated text, as opposed to merely marking whole outputs, is critical for real-world application, as it enables accurate corrections and improved comprehension of where and why a model hallucinate.

\begin{figure}
    \centering
    \includegraphics[width=1\linewidth]{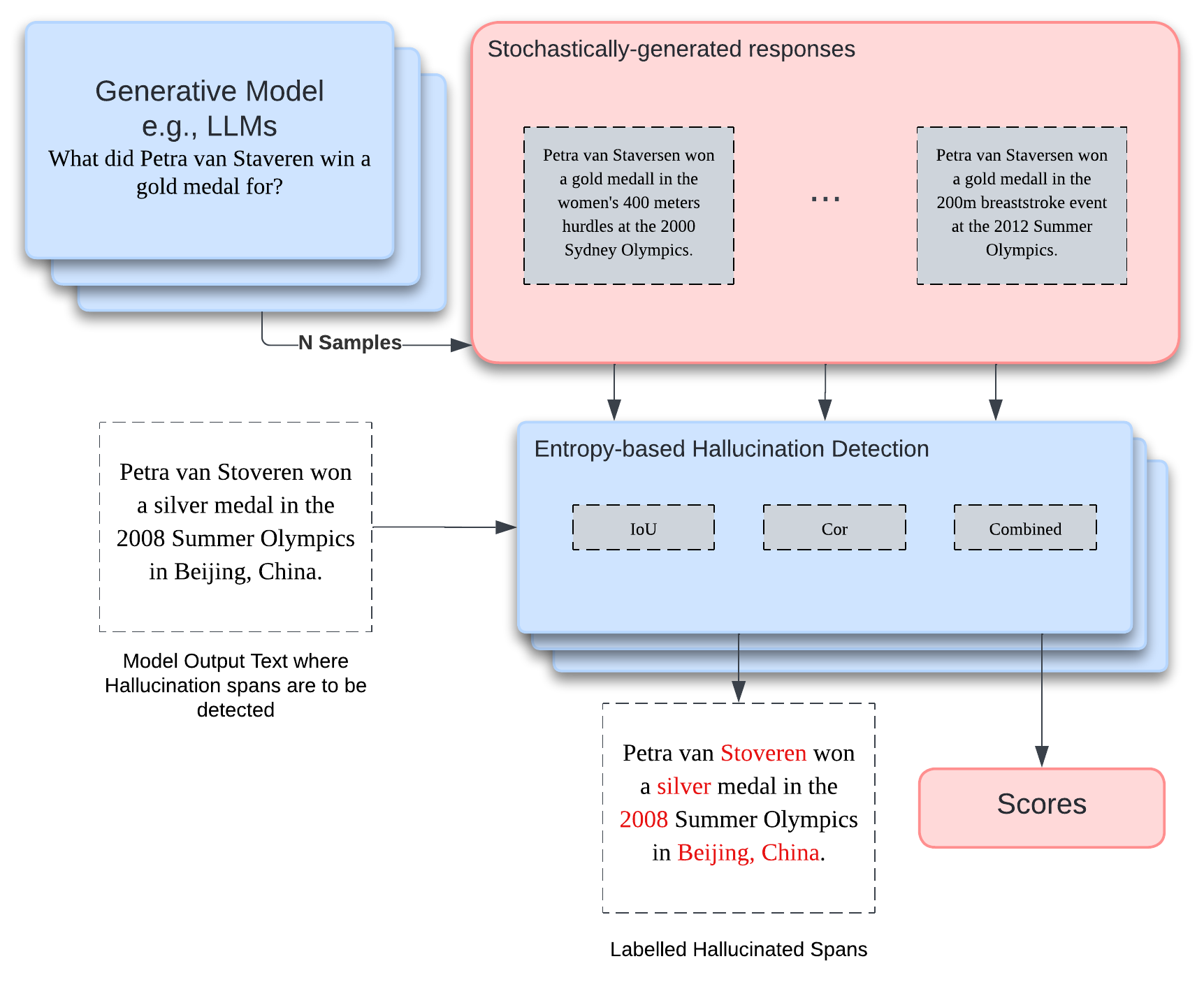}
    \caption{Architecture Diagram describing proposed method for detecting hallucination spans.\citep{selfcheckgpt}}
    \label{fig:enter-label}
\end{figure}

In this paper, we describe an LLM-uncertainty based method for Hallucination span detection. Our hypothesis builds upon \citet{selfcheckgpt} that if an LLM is certain of a given concept, stochastically-sampled responses are likely to be similar and contain consistent facts. However, for hallucinated facts, these sampled responses are likely to diverge and contradict one another. We utilize entropy information to identify the precise spans of hallucinated text using sampled responses \citep{hallucinationUncertainity}, allowing us to effectively identify inconsistencies that signal hallucination.

Our approach works well in zero-resource and black-box environments without any extra training. In addition, since our approach is language-independent, it works equally well in a variety of languages. Our model ranks 18th on average among over 40 submissions, achieving its best rank of 10th in Chinese (Mandarin).\footnote{\url{https://mushroomeval.pythonanywhere.com/submission/}}

\section{Related Work}

The problem of hallucination detection in Large Language Models (LLMs) has been a focus of much attention recently. Hallucinations are defined as cases when LLMs produce outputs that sound plausible but are factually false or unsupported, compromising their validity for real-world usage. \citet{farquhar2024detecting} proposed a technique employing semantic entropy to identify such confabulations through uncertainty estimation in the semantic space of model outputs. This method calculates uncertainty at the meaning level as opposed to actual word sequences and allows for recognizing arbitrary and poor-quality generations for different datasets and tasks without explicit domain knowledge.

Following this, \citet{kossen2024semantic} introduced Semantic Entropy Probes (SEPs), which estimate semantic entropy directly from one generation's hidden states. SEPs are efficient in computation, avoiding repeated model samplings at inference time. Their experiments showed that SEPs have high performance in hallucination detection and generalize well to out-of-distribution test sets, indicating that model hidden states contain semantic uncertainty relevant to hallucinations.

In parallel, \citet{selfcheckgpt} introduced SelfCheckGPT, a zero-resource black-box method for fact-checking LLM responses independent of external databases. The technique exploits the consistency of stochastically generated responses by assuming that when an LLM has knowledge about a concept, its sampled responses will be consistent and similar in content while hallucinated facts result in diverse and contradictory responses. Their results show that SelfCheckGPT efficiently identifies non-factual sentences and evaluates the factuality of passages, providing an efficient solution for situations where model internals are not available.

These studies together highlight the need to create effective and efficient techniques for hallucination detection in LLMs. Methods based on semantic entropy, model hidden states, and response consistency provide promising directions for improving the reliability of LLM outputs in different applications.

\section{Task Description}

Mu-SHROOM \footnote{\url{https://helsinki-nlp.github.io/shroom/}} (Multilingual Shared-task on Hallucinations and Related Observable Overgeneration  Mistakes) focuses on detecting hallucinated spans in text output from instruction-tuned LLMs. The task includes 14 languages: Arabic (Modern standard), Basque, Catalan, Chinese (Mandarin), Czech, English, Farsi, Finnish, French, German, Hindi, Italian, Spanish, and Swedish. \citep{vazquez-etal-2025-mu-shroom}

Evaluation is conducted separately for each language and is based on the following two character-level metrics:
\begin{itemize}
    \item \textbf{Intersection-over-Union (IoU):} Measures the overlap between predicted and reference hallucination spans.
    \[
    \text{IoU} = \frac{|P \cap G|}{|P \cup G|}
    \]
    where \( P \) is the set of predicted hallucination characters and \( G \) is the set of gold reference hallucination characters.

    \item \textbf{Probability Correlation (Cor):} Evaluates how well the predicted hallucination probabilities match empirical annotator probabilities.
    \[
    \rho = \text{corr}(\hat{p}, p)
    \]
    where \( \hat{p} \) are the predicted probabilities and \( p \) are the human-annotated probabilities.
\end{itemize}

\begin{table}[ht]
    \centering
    \renewcommand{\arraystretch}{1.0} % Adjust row height for better readability
    \small
    \begin{tabular}{|l|p{3cm}|} % Increased column width to span two columns
        \hline
        \textbf{Field} & \textbf{Description} \\ 
        \hline
        \texttt{lang} & Language of the text. \\
        \texttt{model\_input} & Input query provided to the LLM. \\
        \texttt{model\_output\_text} & Generated text from the LLM. \\
        \texttt{hard\_labels} & List of pairs \((s_i, e_i)\) representing hallucination spans (start-inclusive, end-exclusive). \\
        \texttt{soft\_labels} & List of dictionaries, each containing:
        \begin{itemize}
            \item \texttt{start}: Start index of hallucination span.
            \item \texttt{end}: End index of hallucination span.
            \item \texttt{prob}: Probability of the span being a hallucination.
        \end{itemize} \\
        \hline
    \end{tabular}
    \caption{Data fields used from Mu-SHROOM Dataset.} \label{tab:data-format}
\end{table}

Data format is described in Table~\ref{tab:data-format}. The \texttt{hard\_labels} are used for intersection-over-union accuracy, while the \texttt{soft\_labels} are used for correlation evaluation. Table~\ref{tab:data-statistics} shows the number of samples in the task dataset.

\section{Methodology}

In this section we describe our methodology for detecting hallucination spans. Given generated text $\mathcal{G}$ and stochastically-sampled responses $\mathcal{S} = {s_1', s_2', ..., s_n'}$ from models, our method predicts hallicination spans as follows:

% \subsection{Sliding Window-Based Text Segmentation}
Given a generated text $\mathcal{G}$, we segment it into overlapping spans using a sliding window approach. Each span $s_i$ is extracted using a window size $w$ and stride $t$ such that:
\begin{equation}
s_i = \mathcal{G}[(i - 1)t:(i - 1)t+w]
\end{equation}
for all valid indices $i$ with step size $t$. This ensures each part of the text is analyzed with sufficient context.

% \subsection{Matching Generated Spans with Sampled Responses}
For each span $s_i$, we retrieve the most similar spans from a set of sampled responses $\mathcal{S} = {s_1', s_2', ..., s_n'}$ using a lexical matching function based on sequence similarity. The matching spans $\mathcal{M}_i$ are defined as:
\begin{equation}
\mathcal{M}_i = {s_j' \in \mathcal{S} \mid \text{Similarity}(s_i, s_j') > \tau}
\end{equation}
where $\tau$ is a threshold for similarity.

% \subsection{Entropy-Based Hallucination Scoring}
We compute the hallucination score for each span $s_i$ using a combination of semantic entropy, lexical entropy, and frequency-based scoring.

\paragraph{Semantic Entropy}
To measure semantic inconsistency, we compute cosine similarity between the span $s_i$ and each matched span $s_j'$, using a pre-trained sentence embedding model:
\begin{equation}
\text{sim}(s_i, s_j') = \frac{E(s_i) \cdot E(s_j')}{|E(s_i)| |E(s_j')|}
\end{equation}
where $E(s)$ denotes the embedding representation of span $s$.
The probability distribution over similarities is given by:
\begin{equation}
P(s_j' \mid s_i) = \frac{e^{\text{sim}(s_i, s_j')}}{\sum_{k} e^{\text{sim}(s_i, s_k')}}
\end{equation}
The semantic entropy is then computed as:
\begin{equation}
H_s(s_i) = - \sum_{s_j' \in \mathcal{M}_i} P(s_j' \mid s_i) \log P(s_j' \mid s_i)
\end{equation}
Higher entropy values indicate greater semantic inconsistency.

\paragraph{Lexical Entropy}
To measure lexical variability, we compute the Shannon entropy over the frequency distribution of matched spans:
\begin{equation}
H_l(s_i) = - \sum_{s_j' \in \mathcal{M}_i} p(s_j') \log p(s_j')
\end{equation}
where $p(s_j')$ is the probability of span $s_j'$ appearing in the matched set $\mathcal{M}_i$.

\paragraph{Frequency Score}
The frequency-based confidence score is computed as:
\begin{equation}
F(s_i) = 1 - \frac{|\mathcal{M}_i|}{|\mathcal{S}|}
\end{equation}
where a lower $|\mathcal{M}_i|$ suggests fewer matches and a higher likelihood of hallucination.

% \subsection{Final Hallucination Score}
The final hallucination score for each span $s_i$ is computed as a weighted sum:
\begin{equation}
S_h(s_i) = \alpha H_s(s_i) + \beta H_l(s_i) + \gamma F(s_i)
\end{equation}

where $\alpha, \beta, \gamma$ are hyperparameters controlling the contribution of each component. For our submission, we heuristically choose $\alpha$ = 0.4, $\beta$ = 0.4 and $\gamma$ = 0.2. We plan to tune these parameters in our future work.

% \subsection{Linguistic Boundary Refinement}
To ensure hallucination spans align with meaningful text units, we refine span boundaries using:
\begin{itemize}
\item \textbf{Token boundaries}: Adjusting span edges to align with word boundaries.
\item \textbf{Phrase boundaries}: Ensuring spans do not split meaningful phrases.
\item \textbf{Named entity boundaries}: Avoiding incorrect segmentation of entity names.
\end{itemize}
The refined spans are selected by maximizing the entropy gradient at span boundaries.

% \subsection{Merging Overlapping Spans}
Detected hallucination spans that overlap significantly are merged into a single span with an updated score:
\begin{equation}
S_h'(s) = \frac{\sum_{i \in \mathcal{O}} S_h(s_i) \cdot |s_i|}{\sum_{i \in \mathcal{O}} |s_i|}
\end{equation}
where $\mathcal{O}$ is the set of overlapping spans.

% \subsection{Final Output}
The final output is a set of hallucination spans $\mathcal{H}$:
\begin{equation}
\mathcal{H} = {(s_i, S_h(s_i)) \mid S_h(s_i) > \lambda }
\end{equation}
where $\lambda$ is a threshold for hallucination detection.

\section{Experiments}

\subsection{Models}

Our experiments utilize Llama-3.2-3B-Instruct model \citep{llama3herdmodels}, a 3 billion parameter instruction-tuned language model. We generate responses using a temperature of 0.1 to maintain relatively deterministic outputs while allowing for some diversity, along with top-p sampling (nucleus sampling) set to 0.9 and top-k sampling with k=50. To avoid repetitive patterns of text, we use a 3-gram repetition penalty. We produce 20 candidate responses with a maximum of 64 tokens per input query. The model is executed in mixed-precision using FP16 to save memory, with memory consumption limited to 6GB GPU memory and 8GB CPU memory via gradient offloading.

\subsection{Hyperparameter Tuning}

Considering the presence of various hyperparameters in our methodology, we perform extensive hyperparameter tuning on validation split for each language. We observe that, while many languages have same set of hyperparameters performing the best on evaluation, there exist few languages where notable differences exist. We summarize our hyperparameters choice in Table~\ref{tab:language_parameters}

\begin{table}[h]
    \centering
    \begin{tabular}{lccccc}
        \toprule
        Language & $w$ & $t$ & $\lambda$ & MSL & BT \\
        \midrule
        arabic  & 4 & 2 & 0.6 & 3 & 0.3 \\
        german  & 4 & 2 & 0.6 & 3 & 0.3 \\
        english & 5 & 3 & 0.5 & 3 & 0.3 \\
        spanish & 4 & 2 & 0.6 & 3 & 0.3 \\
        finnish & 4 & 3 & 0.6 & 3 & 0.3 \\
        french  & 4 & 2 & 0.6 & 3 & 0.3 \\
        hindi   & 5 & 2 & 0.6 & 3 & 0.3 \\
        italian & 4 & 2 & 0.7 & 3 & 0.3 \\
        sweden  & 4 & 2 & 0.5 & 3 & 0.3 \\
        chinese & 7 & 3 & 0.6 & 3 & 0.3 \\
        \bottomrule
    \end{tabular}
    \caption{Hyperparameters choosen for different languages. Notations include $w$: Window Size, $t$: Stride, $\lambda$: Entropy Threshold, MSL: Minimum Span Length, BT: Boundary Threshold}
    \label{tab:language_parameters}
\end{table}

\section{Results and Analysis}

% insert results table here
\begin{table*}[t]
    \centering
    \small
    \renewcommand{\arraystretch}{1.2}
    \setlength{\tabcolsep}{3pt} % Adjust column spacing
    
    % Row 1
    \begin{tabular}{l c c c c c c c c c c}
        \toprule
        Language & \multicolumn{2}{c}{Arabic} & \multicolumn{2}{c}{Catalan} & \multicolumn{2}{c}{Czech} & \multicolumn{2}{c}{German} & \multicolumn{2}{c}{English} \\
        System & IoU & Cor & IoU & Cor & IoU & Cor & IoU & Cor & IoU & Cor \\
        \midrule
        \textit{Baseline (neural)} & 0.0418 & 0.119 & 0.0524 & 0.0645 & 0.0957 & 0.0533 & 0.0318 & 0.1073 & 0.031 & 0.119 \\
        \textit{Baseline (mark none)} & 0.0467 & 0.0067 & 0.08 & 0.06 & 0.13 & 0.1 & 0.0267 & 0.0133 & 0.0325 & 0 \\
        \textit{Baseline (mark all)}  & 0.3614 & 0.0067 & 0.2423 & 0.06 & 0.2632 & 0.1 & 0.3451 & 0.0133 & 0.3489 & 0 \\
        Our Submission & \textbf{0.3631} & \textbf{0.2499} & \textbf{0.3161} & \textbf{0.3377} & \textbf{0.2895} & \textbf{0.2423} & \textbf{0.3651} & \textbf{0.2199} & \textbf{0.366} & \textbf{0.2104} \\
        \midrule
    \end{tabular}

    % Row 2
    \begin{tabular}{l c c c c c c c c c c}
        Language & \multicolumn{2}{c}{Spanish} & \multicolumn{2}{c}{Basque} & \multicolumn{2}{c}{Farsi} & \multicolumn{2}{c}{Finnish} & \multicolumn{2}{c}{French} \\
        System & IoU & Cor & IoU & Cor & IoU & Cor & IoU & Cor & IoU & Cor \\
        \midrule
        \textit{Baseline (neural)} & 0.0724 & 0.0359 & 0.0208 & 0.1004 & 0.0001 & 0.1078 & 0.0042 & 0.0924 & 0.0022 & 0.0208 \\
        \textit{Baseline (mark none)} & 0.0855 & 0.0132 & 0.0101 & 0 & 0 & 0.01 & 0 & 0 & 0 & 0 \\
        \textit{Baseline (mark all)}  & 0.1853 & 0.0132 & 0.3671 & 0 & 0.2028 & 0.01 & \textbf{0.4857} & 0 & 0.4543 & 0 \\
        Our Submission & \textbf{0.2131} & \textbf{0.2335} & \textbf{0.4193} & \textbf{0.3525} & \textbf{0.3132} & \textbf{0.357} & 0.4554 & \textbf{0.3323} & \textbf{0.4651} & \textbf{0.2756} \\
        \midrule
    \end{tabular}

    % Row 3
    \begin{tabular}{l c c c c c c c c}
        Language & \multicolumn{2}{c}{Hindi} & \multicolumn{2}{c}{Italian} & \multicolumn{2}{c}{Swedish} & \multicolumn{2}{c}{Chinese} \\
        System & IoU & Cor & IoU & Cor & IoU & Cor & IoU & Cor \\
        \midrule
        \textit{Baseline (neural)} & 0.0029 & 0.1429 & 0.0104 & 0.08 & 0.0308 & 0.0968 & 0.0236 & 0.0884 \\
        \textit{Baseline (mark none)} & 0 & 0 & 0 & 0 & 0.0204 & 0.0136 & 0.02 & 0 \\
        \textit{Baseline (mark all)}  & 0.2711 & 0 & 0.2826 & 0 & \textbf{0.5373} & 0.0136 & \textbf{0.4772} & 0 \\
        Our Submission & \textbf{0.3598} & \textbf{0.3508} & \textbf{0.4009} & \textbf{0.386} & 0.3967 & \textbf{0.217} & 0.4703 & \textbf{0.1601} \\
        \bottomrule
    \end{tabular}

    \caption{Performance comparison across different languages. IoU ($\protect\Uparrow$) : Intersection over Union. Cor ($\protect\Uparrow$) : Correlation. \textit{Baseline (neural)} represents the baseline provided in participant kit, while \textit{Baseline (mark none)} and \textit{Baseline (mark all)} represents no characters labelled hallucinated and all characters labelled as hallucinated respecitively. $\protect\Uparrow$ denotes higher is better.}
    \label{tab:performance}

\end{table*}

Our submission demonstrated consistent performance across multiple languages as shown in Table~\ref{tab:performance}, achieving similar Intersection over Union (IoU) and Correlation (Cor) scores across various languages. The system performed particularly well in Basque (IoU: 0.4193, Cor: 0.3525), Finnish (IoU: 0.4554, Cor: 0.3323), Italian (IoU: 0.4009, Cor: 0.386) and Hindi (IoU: 0.3598, Cor: 0.3508), indicating its effectiveness in identifying and handling hallucinated text. Similarly, for languages such as English (IoU: 0.3466, Cor: 0.2104), German (IoU: 0.3651, Cor: 0.2199), and Chinese (IoU: 0.4703, Cor: 0.1601), the system maintained consistent performance, demonstrating its adaptability to different linguistic structures.

The findings reveal that our model is aptly suitable for detecting hallucinations for a wide variety of languages that possess intricate morphological and syntactic features. The high correlation scores across numerous languages confirm that our system makes good predictions which correlate well with ground truth annotation. Further, the high IoU values verify its capacity for good localization of hallucinated text, which enables it to be a trustworthy model in addressing the problems of hallucinations in multilingual environments.

\subsection{Error Analysis}

Table~\ref{tab:hallucination} reports a sample data point from test split, where our model's prediction successfully detects the hallucination span. But, it also labels other spans as hallucinated due to noise in generated responses. This behavior of false positives poses significant challenge and it must be handled. We plan to pinpoint why this happens and potentially fix this in our future work.

% \begin{table}[h]
%     \centering
%     \resizebox{\columnwidth}{!}{%
%     \begin{tabular}{@{}p{3cm}p{10cm}@{}}
%         \toprule
%         \textbf{Input Text} & Chi ha doppiato in italiano l'attrice Catherine McCormack nel film Il sarto di Panama? \\ 
%         \midrule
%         \textbf{Ground Truth} & L'attrice Catherine McCormack è stata doppiata in italiano da \textcolor{red}{Elisa Di Stefano} nel film "Il sarto di Panama” \\ 
%         \midrule
%         \textbf{Predicted} & L'attr\textcolor{red}{ice} Catherine McCormack è stata doppiata in italiano da \textcolor{red}{Elisa Di Stefano} nel film "Il sarto\textcolor{red}{ di } \textcolor{red}{Panama}” \\ 
%         \bottomrule
%     \end{tabular}%
%     }
%     \caption{Hallucinated spans highlighted in \textcolor{red}{red} for a sample datapoint in Italian.}
%     \label{tab:hallucination}
% \end{table}

\begin{table*}[h]
    \centering
    \scriptsize
    \resizebox{\textwidth}{!}{%
    \begin{tabular}{@{}lp{3cm}p{6cm}p{6cm}@{}}
        \toprule
        \textbf{Language} & \textbf{Input Text} & \textbf{Ground Truth} & \textbf{Predicted} \\ 
        \midrule
        Italian & Chi ha doppiato in italiano l'attrice Catherine McCormack nel film Il sarto di Panama? & 
        L'attrice Catherine McCormack è stata doppiata in italiano da \textcolor{red}{\underline{Elisa Di Stefano}} nel film "Il sarto di Panama" & 
        L'attr\textcolor{red}{\underline{ice}} Catherine McCormack è stata doppiata in italiano da \textcolor{red}{\underline{Elisa Di Stefano}} nel film "Il sarto\textcolor{red}{\underline{ di }} \textcolor{red}{\underline{Panama}}" \\ 
        \midrule
        English & When did Cape Verde first participate in the Summer Olympics? & Cape Verde participated in its first Summer Olympic games in \textcolor{red}{\underline{2000}} in \textcolor{red}{\underline{Sydney, Australia}}. & Cape Verde \textcolor{red}{\underline{participated in its first Summer Olympic games in} \underline{2000 in Sydney, Australia.}} \\
        \midrule
        Swedish & Vad heter slottsruinen i Pennard? & Den kallades \textcolor{red}{\underline{vid tiden för sitt största riktigt, Castle of the} \underline{Dromon}}. Det är en liten ruin på ett berg nära \textcolor{red}{\underline{stranden av St Brides Bay}}. & Den kallades vid \textcolor{red}{\underline{tiden för sitt största riktigt, Castle of the} \underline{Dromon}}. Det är en liten ruin på ett berg \textcolor{red}{\underline{nära stranden}} av St Brides \textcolor{red}{\underline{Bay}}.\\
        \bottomrule
    \end{tabular}%
    }
    \caption{Hallucinated spans highlighted in \textcolor{red}{\underline{red}} for a sample datapoints in some langauges. Predicted refer to the predicted hallucinated spans by our system.}
    \label{tab:hallucination}
\end{table*}

\section{Conclusion}

In this paper, we utilized an LLM-uncertainty-based method for hallucination span detection which works equally well in multiple languages. By using entropy-based uncertainty measures from sample responses, our approach accurately detects hallucinated spans without the need for further training. Our model performed competitively in various languages, ranking highly in Basque, Italian, and Hindi. The experiments emphasize the strength of our method, as they show its effectiveness in coping with varied linguistic forms and in yielding precise hallucination span detection. Our error analysis also informs on typical failure instances, presenting potential for additional refinements.

Although our approach is strong, it has limitations, specifically in exploiting supervised learning to achieve better span prediction. Our future research might consider fine-tuning over accessible training data in order to make performance even better while keeping our zero-resource model flexible. More context and fact-based verification methods can be incorporated to improve hallucination detection even further. With LLMs still evolving, creating scalable and accurate methods of hallucination detection remains a critical step to maintain the integrity of AI-produced text across real-world use cases.

\section*{Limitations}

Our method does not employ supervised learning for predicting the exact spans. Under-utilization of training splits of the task is a major drawback of our system. Utilizing the training split for any kind of supervised learning could potentially improve the performance. Moreover, failing to incorporate contextual and factual verification techniques poses a major challenge to our approach.  

\section*{Acknowledgments}

We would like to thank Mu-SHROOM shared task organizers, Raúl Vázquez, Timothee Mickus, and their team, for their effort and commitment to organizing this task.

\bibliography{custom}

\appendix

\section{Mu-SHROOM Dataset Statistics}
\label{sec:appendix}

\begin{table}[h]
    \centering
    \scriptsize
    % \small
    \begin{tabular}{l|cc}
        \toprule
        Language & Validation & Test \\
        \midrule
        ar & 50 & 150 \\
        ca & - & 100 \\
        cs & - & 100 \\
        de & 50 & 150 \\
        en & 50 & 154 \\
        es & 50 & 152 \\
        eu & - & 100 \\
        fa & - & 100 \\
        fi & 50 & 150 \\
        fr & 50 & 150 \\
        hi & 50 & 150 \\
        it & 50 & 150 \\
        sv & 50 & 150 \\
        zh & 50 & 150 \\
        \bottomrule
    \end{tabular}
    \caption{Number of Samples in Validation and Test data in Mu-SHROOM. For Hyperparameter Tuning, we considered validation split for languages containing validation data points. For others, we heuristically approximate the parameters.}
    \label{tab:data-statistics}
\end{table}

\end{document}